\documentclass[AMA,Times2COL]{WileyNJDv5} 
\pdfoutput=1

\articletype{Article Type}%

\received{Date Month Year}
\revised{Date Month Year}
\accepted{Date Month Year}
\journal{Journal}
\volume{00}
\copyyear{2023}
\startpage{1}

\usepackage{xcolor}
\usepackage[table,xcdraw]{xcolor}
\usepackage{array}
\begin{document}

\title{FactCheck: Feasibility-aware Long-term Action Anticipation with Multi-agent Collaboration}
\author[1]{Rui Cao}
\author[1]{Jiannong Cao}
\author[2]{Bo Yuan}
\author[1]{Zhiyuan Wen}
\author[1]{Mingjin Zhang}



\authormark{CAO \textsc{et al.}}
\titlemark{FactCheck: Feasibility-aware Long-term Action Anticipation with Multi-agent Collaboration}

\address[1]{\orgdiv{Department of Computing}, \orgname{The Hong Kong Polytechnic University}, \orgaddress{\state{Hong Kong SAR}, \country{China}}}

\address[2]{\orgdiv{JIUTIAN Research}, \orgname{China Mobile}, \orgaddress{\state{Beijing}, \country{China}}}

\corres{Corresponding author Zhiyuan Wen, The Hong Kong Polytechnic University, 11 Yuk Choi Road, Hung Hom, Kowloon, Hong Kong SAR, China, \email{zhiyuan.wen@polyu.edu.hk}\\
Corresponding author Bo Yuan, JIUTIAN Research, China Mobile, 28 Financial Street, Xicheng District, Beijing, China, \email{yuanbo@cmjt.chinamobile.com
}}




\abstract[Abstract]{Long-term action anticipation (LTA) aims to predict an ordered sequence of future verb-noun actions from a partially observed video. While this task serves as the foundation for embodied intelligence, anticipating physically feasible long-term actions remains a critical challenge. Existing methods, which operate in an open-loop manner, often hallucinate non-existent objects, violate object affordances, or disregard object states, as they lack explicit mechanisms to verify action feasibility against the physical environment. To address this, we propose FactCheck, a novel multi-agent collaboration framework that improves feasibility through a closed-loop "Observe-Plan-Verify" mechanism. FactCheck decomposes the complex LTA task into specialized roles: 
an Observer that recognizes historical actions from video observations and constructs a dual-form structured memory, comprising a History Action Abstract that captures high-level human intentions and environmental status, and a History Action Graph that encodes object states and temporal dependencies;
a Planner that generates draft future actions conditioned on both low-level historical actions and high-level History Action Abstract; and a Verifier that rigorously validates the draft against the History Action Graph and refines infeasible actions. Extensive experiments on the EPIC-Kitchens-55 and EGTEA Gaze+ benchmarks demonstrate that FactCheck consistently outperforms state-of-the-art methods. Our work establishes a new paradigm for feasibility-aware long-term action anticipation, effectively closing the loop of action recognition, action prediction and action verification.}


\jnlcitation{\cname{%
\author{Taylor M.},
\author{Lauritzen P},
\author{Erath C}, and
\author{Mittal R}}.
\ctitle{On simplifying ‘incremental remap’-based transport schemes.} \cjournal{\it J Comput Phys.} \cvol{2021;00(00):1--18}.}

\maketitle



\section{Introduction}
\label{sec1:introduction}
Long-term action anticipation (LTA) is a critical task for embodied intelligence~\cite{nawhal2022rethinking, chang2020procedure}, as it enables agents to learn complex skills from human demonstrations~\cite{verghese2025skills, li2025learning} and plays a key role in providing proactive assistance during human-robot collaboration~\cite{de2025pace, hoffman2024inferring}.
Specifically, given an egocentric video sequence from a human, the objective of LTA is to predict a series of future actions performed by this camera wearer.
Each action is typically formatted as a \textit{verb-noun} pair, where the noun identifies the interactive object and the verb describes the specific motion~\cite{hussein2019videograph, zhang2024object}.

However, long-term action anticipation (LTA) is also challenging, as it demands both \textit{historical action recognition} to understand the visual context and \textit{future action prediction} to forecast the semantic verb-noun action pairs~\cite{carreira2017quo, abu2018will, gong2022future}.
Another fundamental challenge of LTA lies in its extended prediction horizon.
Different from \textit{short-term action anticipation}, which typically focuses on predicting the immediate next action in a few seconds~\cite{fernando2021anticipating, roy2024interaction, cao2024vs}, \textit{long-term action anticipation} aims to predict a sequence of future actions over a much broader horizon, typically spanning several minutes~\cite{grauman2022ego4d, zhao2024antgpt, kim2024palm}. While LTA serves as the cornerstone of embodied intelligence, this task remains challenging.
The difficulty of LTA stems not only from the complex temporal dynamics of long-term actions but also from the latent human intentions underlying these actions.

LTA remains particularly challenging due to three aspects. First, it demands both historical action recognition to understand the visual context and future action prediction to forecast semantic verb-noun action pairs~\cite{carreira2017quo, abu2018will, gong2022future}, requiring methods to excel at both ends. Second, unlike short-term action anticipation that predicts the immediate next action within a few seconds~\cite{fernando2021anticipating, roy2024interaction, cao2024vs}, LTA forecasts action sequences over a much broader horizon spanning several minutes~\cite{grauman2022ego4d, zhao2024antgpt, kim2024palm}. This leads to growing temporal complexity and compounding prediction errors. Third, the actions in LTA are driven by implicit human intentions that are difficult to directly observe, yet critically determine the course of future actions.

\begin{figure*}[t]
    \centering
    \vspace{-8pt} 
    \includegraphics[width=0.75\linewidth]{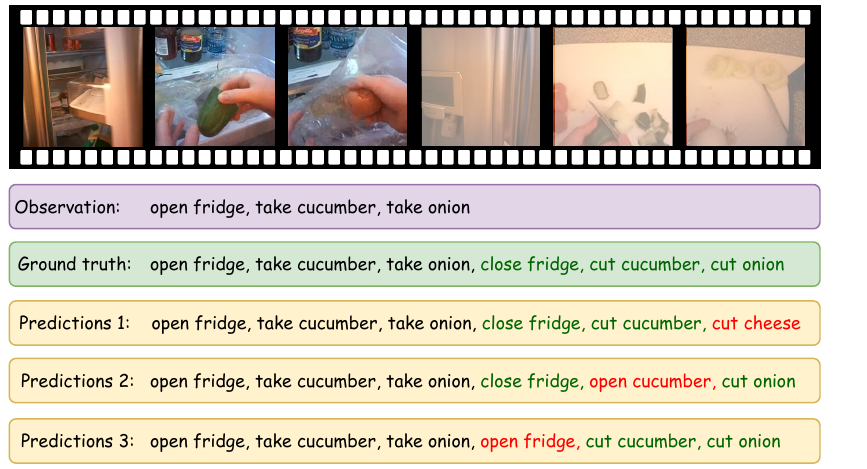}
    \vspace{-1mm}
    \caption{Illustration of infeasible action predictions in the existing Long-Term Action Anticipation (LTA) models. Given the video observation above, current open-loop models often generate predictions that are temporally consistent and semantically coherent, but physically unexecutable. Prediction 1 violates \textit{Object Availability} by hallucinating a non-existent object (``cut cheese''). Prediction 2 violates \textit{Affordance Validity} by predicting an interaction unsupported by the object category (``open cucumber''). Prediction 3 violates \textit{State Consistency} by attempting to ``open'' a fridge that is already in an open state. These failures highlight the critical limitation of neglecting action feasibility in LTA.}
    \label{fig:limitation}
    \vspace{-5mm}
\end{figure*}

To address these difficulties, existing approaches have primarily evolved along two distinct paradigms:
(1) Data-driven temporal modeling methods: Early research predominantly formulates LTA as a sequence modeling task and captures the dependencies between observed visual data to future actions through learned temporal representations. Specifically, architectures such as recurrent neural networks (RNNs)~\cite{shi2018action, sener2020temporal, roy2024predicting}, long short-term memory (LSTMs)~\cite{furnari2020rolling, osman2021slowfast}, and 
transformers~\cite{girdhar2021anticipative, gong2022future, wu2022memvit} are employed to encode the observed video frames and predict future actions. While effective on specific benchmarks, these data-driven methods heavily rely on domain-specific training data. This reliance limits their ability to generalize to long-tail classes or unseen scenarios and neglects underlying human intention~\cite{kim2024palm}.
(2) Large Language Model (LLM)-driven agentic reasoning methods: To mitigate data dependency and reason human intention, recent advancements instantiate agents based on pretrained LLMs to generate future actions adaptively~\cite{kim2024palm, zhao2024antgpt, mittal2024can, wang2025multimodal}. These methods first encode video context into textual action descriptions, the LLM agents then generate future action sequences only requiring lightweight fine-tuning or in-context learning (ICL). Additionally, by leveraging the rich world knowledge, LLM agents are capable of inferring the action narrations or latent human intentions behind observed behaviors. This high-level reasoning ability enables the LLM agent to generate semantically coherent action sequences.

Despite their respective strengths in temporal consistency~\cite{shi2018action, furnari2020rolling, girdhar2021anticipative} or semantic coherence~\cite{kim2024palm, zhao2024antgpt, mittal2024can, wang2025multimodal}, both paradigms share a fundamental limitation: they operate in an open-loop prediction manner, largely ignoring the physical feasibility of predicted actions in the actual environment. As a result, existing approaches frequently generate actions that are unrealistic or impossible to execute. As illustrated in Figure~\ref{fig:limitation}, previous methods may exhibit three typical types of feasibility failures: (1) Hallucinating non-existent objects: Prediction 1 attempts to "cut cheese", ignoring that no cheese is present in the scene. (2) Violating object affordance: Prediction 2 predicts "open cucumber", failing to recognize that a cucumber does not possess the affordance to be opened. (3) Disregarding object state: Prediction 3 attempts to "open" a "fridge" that is already open, conflicting with the object's current physical status.

Those feasibility failures stem from the lack of mechanisms to verify whether the generated actions meet the physical requirements. More specifically, 
(1) \textit{Object Availability}: The interactive object must be visually observable in the current environment.
(2) \textit{Affordance Validity}: The predicted action must be compatible with the inherent physical properties of the object category (e.g., openable vs. cuttable).
(3) \textit{State Consistency}: The action must be valid given the object's current dynamic state (e.g., an object must be closed before it can be opened).
Existing models may violate these constraints, leaving the problem of infeasible action generation unresolved.

However, implementing an effective feasibility verification mechanism is non-trivial and introduces several research challenges. 
The first challenge lies in the egocentric video understanding: 
\textit{How can we extract the global memory of the environment from unstructured egocentric videos?} 
Egocentric videos suffer from severe occlusion and dynamic viewpoint changes, where objects frequently vanish from the current view.
Thus, single-agent approaches often suffer from contextual forgetting, struggling to maintain a global memory of the environment from the video observation.
The second challenge lies in the action prediction: \textit{How can we predict future actions that are consistent with the global memory?} Existing LLM-based approaches typically rely on either fine-grained historical action narrations~\cite{kim2024palm} or high-level human intentions~\cite{zhao2024antgpt} to predict future actions. As a result, the generated actions may lack either step-by-step coherence or high-level goal consistency, therefore neglecting the feasibility of the future actions. 
The third challenge concerns the action verification: \textit{How can we validate the feasibility of action predictions?} Most existing methods~\cite{gong2022future, cao2025vision} predict actions without any explicit verification, overlooking the feasibility constraints of the environment.
In summary, existing single-agent approaches struggle with contextual forgetting and lack the mechanism to explicitly verify physical states.
Building upon the above challenges, there is a need for a collaborative mechanism to conduct long-term action anticipation through action recognition, action prediction, and action verification under feasibility constraints.

In this paper, we propose FactCheck, a multi-agent collaboration framework designed for feasibility-aware long-term action anticipation (LTA). To mitigate the contextual forgetting inherent in single-agent solutions and enhance the ability to verify action predictions, FactCheck decomposes the complex LTA task into specialized subtasks, each handled by a dedicated agent within a collaborative \texttt{"Observe-Plan-Verify"} mechanism. To address the first challenge of extracting the global memory from the egocentric video, we introduce an \texttt{Observer} agent as the perception and memory unit. This agent transforms historical observations into a dual-form structured memory. Specifically, it infers a History Action Abstract to capture high-level human intentions and low-level environmental status. Moreover, the Observer agent generates a History Action Graph to provide a topological record of object states and temporal dependencies. 
To tackle the second challenge of action prediction, we propose a \texttt{Planner} agent. By synthesizing the dual-level guidance from the History Action Abstract, the \texttt{Planner} is fine-tuned via QLoRA to maintain high semantic coherence across variable-length sequences, ensuring the predictions align with both the high-level human intentions and low-level environmental status. To resolve the third challenge of action verification, we develop a \texttt{Verifier} agent to close the loop between action prediction and the physical environment. Unlike open-loop single-agent methods that lack an explicit action verification, the \texttt{Verifier} records the topological structure of object states and temporal dependencies to perform a three-phase feasibility verification through Chain-of-Thought (CoT) reasoning. This specialized agent systematically validates and corrects violations in the draft actions. This verification procedure ensures the predictions are physically executable.
Experiments on the Epic-Kitchen 55~\cite{damen2020epic} and EGTEA Gaze+~\cite{li2021eye} benchmarks demonstrate that FactCheck outperforms existing state-of-the-art (SoTA) methods, achieving performance gains of 3.1\% and 1.1\%, respectively.

Specifically, our contributions in this paper are as follows:
\begin{itemize}
\item We propose FactCheck, a multi-agent collaboration framework that enables feasibility-aware long-term action anticipation (LTA) through a closed-loop "Observe-Plan-Verify" mechanism.
\item We introduce an Observer agent that serves as the perception and memory unit to obtain a History Action Abstract and a History Action Graph from the video observation. 
\item Based on the History Action Abstract, we propose a Planner agent to predict future actions that are consistent with environmental status and human intention. 
\item Based on the History Action Graph, we develop a Verifier agent to explicitly detect infeasible actions and refine them to ensure feasibility-aware LTA.
\item By closing the loop of action recognition, action prediction and action verification, FactCheck outperforms existing SoTA methods in both the Epic-Kitchen 55 and EGTEA Gaze+ benchmarks.
\end{itemize}

\section{Related Works}
\label{sec2}
This section reviews the literature relevant to our work. 
First, we discuss existing paradigms in action anticipation, highlighting the shift from data-driven temporal modeling to LLM-based agentic reasoning. 
Second, we introduce recent advancements in LLM-based multi-agent collaboration, with a specific focus on the memory mechanism and verification mechanism that motivate our framework.

\subsection{Action Anticipation}
Given an observed video segment, action anticipation aims to predict a sequence of future actions, typically formatted as \textit{verb-noun} pairs. Existing works have addressed this challenge across various domains, including both third-person~\cite{abu2018will,chen2022gatehub,rizve2023pivotal} and egocentric~\cite{gong2022future,zhang2024object, zhao2024antgpt} video settings.
However, egocentric action anticipation poses unique challenges due to drastic camera motion, severe object occlusions, and the inherent ambiguity of inferring latent human intentions from a first-person perspective~\cite{wu2022memvit, kim2024palm, mittal2024can}.
Unlike short-term anticipation, which focuses on the immediate next action~
\cite{fernando2021anticipating,ragusa2023stillfast,roy2024interaction, cao2024vs}, long-term action anticipation (LTA) requires modeling complex temporal dependencies and reasoning about the future action sequence over several minutes~\cite{gong2022future}. To address these difficulties, existing LTA approaches have primarily evolved along two distinct paradigms: Data-driven temporal modeling methods and LLM-driven agentic reasoning methods.

\textbf{Data-driven temporal modeling methods.} 
Early approaches treated LTA as a sequence modeling problem, utilizing recurrent neural networks (RNNs)~\cite{shi2018action,sener2020temporal,roy2024predicting}, long short-term memory (LSTMs)~\cite{furnari2020rolling,osman2021slowfast} to encode observed frames and autoregressively decode future actions. Transformer-based architectures have shown stronger capability in modeling long-horizon temporal dependencies.
FUTR~\cite{gong2022future} introduces an end-to-end Future Transformer that leverages global attention over all input frames and output tokens, predicting the future actions in parallel decoding. 
ObjectPrompt~\cite{zhang2024object} argues that objects provide critical cues for future interactions. It enhances the Transformer architecture by incorporating object-centric video representations, using "object prompts" to query pre-trained vision-language models and retrieve task-relevant object information without expensive fine-tuning.
Modeling long-term dependencies often incurs high computational costs due to the need to process long video history.
To address this memory bottleneck, MeMViT~\cite{wu2022memvit} introduces a memory-augmented Multiscale Vision Transformer that processes video in an online manner. EGO-TOPO~\cite{nagarajan2020ego} constructs a topological graph of the environment to organize the video into a series of spatial zones, linking these zones to their functional affordances~\cite{nagarajan2020ego}.
Other methods enhance the temporal modeling via skip-connections~\cite{ke2019time}, message passing neural networks~\cite{tai2022unified}, and generative models~\cite{mascaro2023intention}.

\textbf{LLM-driven agentic reasoning methods.} 
While data-driven methods excel at pattern recognition, the reliance on large datasets limits the model's ability to generalize to less common, long-tail classes and unseen scenarios.
Additionally, data-driven methods typically lack the explicit, semantic-level reasoning capability to interpret human intent. To bridge this gap, recent methods alleviate the dependency on large-scale human action video datasets by reasoning about the high-level human intention of the task goal.
AbstractGoal~\cite{roy2024predicting} attempted to implicitly model the abstract goal via a stochastic RNN. 
However, the learned latent representations lack the explicit semantic reasoning capabilities required to interpret complex human intentions, paving the way for LLM-based agentic reasoning.
AntGPT~\cite{zhao2024antgpt} formulates LTA from a top-down perspective, where an LLM first explicitly infers the actor's high-level goal from observed actions using in-context learning, and then generates the future action sequence to achieve that goal. 
PALM~\cite{kim2024palm} bridges the modality gap by converting visual context into narrative captions via a Vision-Language Model (VLM), feeding these descriptions into an LLM agent to reason about future steps. 
To further refine generation quality, PlausiVL~\cite{mittal2024can} introduces a "plausibility" constraint, fine-tuning the LLM with counterfactual losses to implicitly ensure predicted sequences adhere to real-world temporal logic. 
ICVL~\cite{cao2025vision} explicitly infers behavioral intentions as textual features using a VLM and fuses them with visual embeddings, creating intention-enhanced representations that guide the LLM agent to generate more consistent long-term plans. 
These methods demonstrate that leveraging the commonsense knowledge embedded in LLM agents significantly enhances the semantic coherence and intentional reasoning in action anticipation with high-level human intentions.

However, most existing models perform action anticipation in an open-loop manner, directly generating actions without explicitly verifying their physical feasibility in the current environment. Differently, our work proposes a closed-loop multi-agent framework, ensuring that long-term action anticipation is not only semantically coherent but also physically executable.

\subsection{LLM-based Multi-agent Collaboration}
Although single LLM agents have demonstrated strong reasoning and planning abilities, they still lack an adequate long-context processing capability and suffer from cascading hallucinations. 
To address these limitations, recent research has shifted towards LLM-based multi-agent collaboration systems~\cite{li2024survey,zhao2025codeedu}.
This paradigm tackles complex problems through the collaboration of multiple specialized LLM agents.
Recent approaches have demonstrated improved performance by incorporating different memory and verification mechanisms.

\textbf{Memory mechanism in multi-agent collaboration.}
Serving as the storage and retrieval unit, the memory module equips the agent with the capacity to utilize existing cognitive and experiential knowledge~\cite{li2024survey}. Existing works primarily address two aspects: memory storage and memory retrieval.
On one hand, memory storage is the process of archiving the information perceived and the experiences learned by agents during interactions into natural language~\cite{achiam2023gpt}.
LLM agents also encompass multimodal information, such as visual data~\cite{arefeen2024vita,liu2025hm} and audio signals~\cite{lu2025vidove,dutta2025can}. 
To achieve efficient and flexible information retention, existing methodologies employ improved data storage structures rather than simple logging. For instance, some approaches adopt embedding vectors to represent memory sections and historical dialogues~\cite{zhu2023ghost, qian2024chatdev, lin2023agentsims}, while others generate SQL-like instructions for structured data manipulation~\cite{hu2023chatdb}. 
On the other hand, memory retrieval facilitates dynamic interactions by autonomously extracting valuable information relevant to the current situation from the agent's memory. Park et al.~\cite{park2023generative} emphasize evaluating memory information based on predefined metrics such as recency, relevance, and importance to prioritize the most significant memories for the current context. MemoryBank employs a dual-tower dense retrieval architecture to encode dialogue contexts and memory fragments into vectors for efficient semantic matching~\cite{zhong2024memorybank}. 
To further handle multimodal data, HM-RAG orchestrates specialized retrieval agents to conduct parallel searches across vector, graph, and web-based databases for comprehensive knowledge synthesis~\cite{liu2025hm}.

\textbf{Verification mechanism in multi-agent collaboration.}
Ensuring the reliability of agent behaviors and mitigating hallucinations are critical challenges in multi-agent systems~\cite{li2024survey}. To address these issues, current research predominantly employs two strategies: multi-agent debate (MAD) and dedicated verifier roles.
In the MAD framework, agents verify and refine outputs through iterative arguments to correct each other's errors and reach a more accurate conclusion~\cite{liang2024encouraging, ning2025mad}. 
Similarly, ChatEval constructs a multi-agent referee team to evaluate generated text, significantly improving the consistency with human judgments through collaborative communication~\cite{chan2024chateval}. 
Beyond peer discussion, structural verification introduces specific agents tasked solely with oversight. VeriMAP~\cite{xu2025verification} proposes a verification-aware planning system where a dedicated Verifier module operates alongside Planner and Executor agents to validate subtask outputs before proceeding. In hierarchical systems like MegaAgent~\cite{wang2025megaagent}, a Boss Agent performs system-level monitoring to review the outputs of all agent groups, ensuring consistency and adherence to global constraints.
Thucy~\cite{theologitis2025thucy} employs a specialized Verifier agent that orchestrates SQL experts to verify claims against relational databases, ensuring responses are grounded in structured evidence.

Building upon these advancements, we develop a multi-agent framework that explicitly integrates memory and verification for LTA. Unlike prior open-loop approaches, our system performs through a closed-loop collaboration mechanism, ensuring that anticipated actions are both semantically coherent and physically feasible in dynamic environments.

\section{Problem Formulation}
\label{sec3:problem_formulation}

Long-term Action Anticipation (LTA) is the task of predicting a sequence of future actions based on historical video observations.
Formally, consider an egocentric video observation $\mathcal{O}$ consisting of $M$ action segments. Let $\alpha \in [0, 1]$ denote the observation ratio and $\beta \in [0, 1-\alpha]$ denote the prediction ratio.
The video observation is partitioned into a sequence of segments $\mathcal{O} = \{ (S_k, a_k) \}_{k=1}^{M}$, where each $S_k$ represents a video segment and $a_k = (v_k, n_k)$ is the corresponding action label comprising a verb $v_k \in \mathcal{V}_{verb}$ and a target object noun $n_k \in \mathcal{N}_{obj}$.

Given an observation ratio $\alpha$, the historical video segments are $\mathcal{O}_{hist} = \{ (S_k, a_k) \}_{k=1}^{N}$, where $N = \lfloor \alpha M \rfloor$ is the number of observed segments.
Accordingly, we denote the sequence of recognized historical actions as $\mathcal{A}_{hist} = \{a_1, \dots, a_N\}$.
The objective of LTA is to predict the subsequent sequence of future actions $\mathcal{A}_{fut} = \{a_{N+1}, \dots, a_{N+Z}\}$ spanning a horizon $Z = \lfloor \beta M \rfloor$.

Prior LTA approaches typically formulate this task as maximizing the conditional probability $P(\mathcal{A}_{fut} | \mathcal{O}_{hist})$~\cite{abu2018will,gong2022future,zhao2024antgpt}. However, purely maximizing likelihood often ignores the physical causality of the real world.
Let $\mathcal{N}_{obs}$ denote the set of the objects observed in $\mathcal{O}_{hist}$. 
For each object $n \in \mathcal{N}_{obs}$, let $t^*(n) = \max \{j \le N \mid n_j = n\}$ be the index of its most recent interaction. 
The current environment state $\text{Env}_N$ is defined as the aggregation of the states of all observed objects, determined by their last associated verbs:
\begin{equation}
    \text{Env}_N = \{ (n, \text{Post}(v_{t^*(n)})) \mid n \in \mathcal{N}_{obs} \},
\end{equation}
where $\text{Post}(v)$ denotes the state resulting from verb $v$ (e.g., ``open'' $\to$ ``opened'').

A predicted action $a_j$ (for $j > N$) is considered valid only if it is physically executable given the evolving state derived from the history.
To explicitly model this dependency, we introduce three feasibility guidelines on each predicted action $a_k = (v_k, n_k)$.

\begin{itemize}
    \item \textbf{Object Availability ($\mathcal{C}_{obj}$):} 
    To mitigate object hallucinations, the agent prioritizes objects that are already observed in the current environment.
    We define the available object set $\mathcal{N}_{obs}$ as the set of objects observed in 
    the historical observation $\mathcal{O}_{hist}$. The constraint is defined as:
    \begin{equation}
        \mathcal{C}_{obj}(a_j, \mathcal{O}_{hist}): \quad n_j \in \mathcal{N}_{obs}.
    \end{equation}

    \item \textbf{Affordance validity ($\mathcal{C}_{aff}$):} 
    The action verb $v_j$ is feasible if the target object $n_j$ affords this specific interaction. Let $\text{Aff}(n) \subseteq \mathcal{V}_{verb}$ be the set of valid affordances for object noun $n$. The constraint is:
    \begin{equation}
        \mathcal{C}_{aff}(a_j): \quad v_j \in \text{Aff}(n_j).
    \end{equation}

    \item \textbf{State Consistency ($\mathcal{C}_{sta}$):} 
    The action verb $v_j$ is feasible if the current state of the object $n_j$ satisfies the preconditions required by the action.
    For a predicted action $a_k = (v_k, n_k)$, we retrieve the current state of object $n_k$ based on its most recent interaction time $t^* = \max \{j < k \mid n_j = n_k\}$.
    The constraint ensures that the state resulting from the previous interaction satisfies the precondition of the current verb $v_k$:
    \begin{equation}
        \mathcal{C}_{sta}(a_k, \mathcal{O}_{hist}): \quad \text{Post}(v_{t^*}) \in \text{Pre}(v_k),
    \end{equation}
    where $\text{Pre}(v_k)$ denotes the set of requisite preconditions for verb $v_k$.
    
\end{itemize}

\begin{figure*}[htb]
    \centering
    \vspace{-8pt} 
    \includegraphics[width=0.9\linewidth]{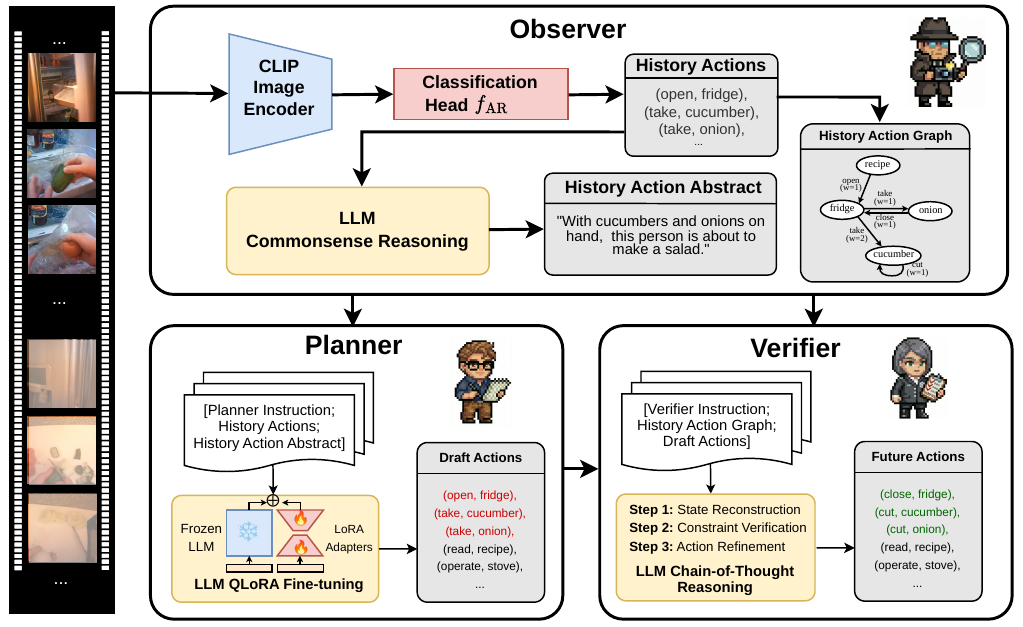}
    \vspace{-1mm}
    \caption{
    \textbf{Overview of the FactCheck Framework.} 
    Our approach achieves feasibility-aware Long-Term Action Anticipation (LTA) through a closed-loop \texttt{“Observe-Plan-Verify”} mechanism:
    (1) The \texttt{Observer Agent} functions as the perception and memory core. It employs an Action Recognition Module to extract historical actions from video frames to obtain the History Action Abstract, and the History Action Graph to encode object states and temporal dependencies.
    (2) The \texttt{Planner Agent} acts as the action prediction unit. Conditioned on the historical actions and the History Action Abstract, it utilizes a QLoRA fine-tuned LLM to generate future draft actions.
    (3) The \texttt{Verifier Agent} serves as the feasibility verification unit. It validates the draft against the History Action Graph using Chain-of-Thought (CoT) reasoning to enforce \textit{Object Availability}, \textit{Affordance Validity}, and \textit{State Consistency}. The identified violations are corrected to ensure the verified actions are physically executable.
    }
    \label{fig:framework}
    \vspace{-5mm}
\end{figure*}

Consequently, we formulate feasibility-aware LTA as a constrained optimization problem. Our goal is to predict the future action sequence $\mathcal{A}^*_{fut}$ that maximizes the joint probability while satisfying the proposed feasibility constraints at each time step.
Formally, the objective is defined as:
\begin{align}
    \label{eq:objective}
    \mathcal{A}^*_{fut} = \operatorname*{arg\,max}_{\mathcal{A}_{fut}} & \sum_{k=N+1}^{N+Z} \log P(a_k \mid \mathcal{O}_{hist}, a_{N+1:k-1}) \\
    \text{s.t.} \quad & \mathcal{C}_{obj}(a_k) \land \mathcal{C}_{aff}(a_k) \land \mathcal{C}_{sta}(a_k) \iff \text{True}, \forall k. \notag
\end{align}

\section{Methodology}
\label{sec4:methodology}

In this section, we present \textbf{FactCheck}, a multi-agent collaboration framework designed to achieve feasibility-aware Long-Term Action Anticipation (LTA). As illustrated in Figure~\ref{fig:framework}, our framework implements a closed-loop \texttt{“Observe-Plan-Verify”} mechanism that iteratively grounds high-level semantic planning in low-level physical constraints. This mechanism orchestrates three specialized agents, each designed to address a core challenge outlined in Section~\ref{sec1:introduction}: (1) The \texttt{Observer} agent serves as the perception and memory unit. It processes raw video observations to recognize history actions, subsequently constructing and maintaining a global History Action Abstract and a structured History Action Graph. (2) The \texttt{Planner} agent acts as the action prediction unit. Conditioned on the extracted historical actions and the global memory, it generates a draft sequence of future actions. This agent leverages a Large Language Model (LLM) that is efficiently fine-tuned via QLoRA to learn task-specific temporal dynamics. (3) The \texttt{Verifier} agent functions as the feasibility verification unit. It rigorously validates each draft action against the History Action Graph using Chain-of-Thought (CoT) reasoning to check for violations of the three feasibility constraints. The infeasible actions are then corrected through a targeted In-Context Learning (ICL) refinement step, ensuring the final predictions are both semantically coherent and physically executable. This collaborative, closed-loop process ensures that long-term action anticipation is feasible in the real-world environment.

\subsection{Observer Agent}
\label{sec4-1:observer}
The Observer agent functions as the perception and memory core of FactCheck. It encodes raw egocentric video streams into structured history action knowledge through three components:
(1) an Action Recognition module that extracts past verb-noun pairs from the observed frames;
(2) a History Action Abstract Reasoning module that infers the underlying task goal and environmental status; and (3) a History Action Graph Construction module that maintains a topological graph of object states and their temporal dependencies. This structured memory forms the essential environmental context for the subsequent planning and verification stages.

\noindent \textbf{Action Recognition Module.}
To initiate the anticipation process, this module functions as the core perception unit, extracting historical actions from the video input $\mathcal{O}_{hist}$. Given the observed video segments $\{S_1, S_2, \dots, S_N\}$, the module produces a sequence of past actions $\{a_1, a_2, \dots, a_N\}$, where each $a_k = (v_k, n_k)$ consists of a verb $v_k$ and a noun $n_k$.
Leveraging on the strong vision-language alignment capability learned from large-scale image-text pairs, we employ the pretrained CLIP-ViT-L/14-336px model~\cite{radford2021learning} for feature extraction. For each video segment $S_k$, we uniformly sample $T$ frames and extract frame-level CLIP image features $\{{o}_1, {o}_2, \dots, {o}_T\}$, where $o_t \in \mathbb{R}^{d{clip}}$. These features are temporally aggregated via average pooling to produce a segment-level representation ${x}_k \in \mathbb{R}^{d{clip}}$.
This open-vocabulary paradigm enables the agent to capture semantic-aware image embeddings for further action recognition. These features ${x}_k$ are then classified into the corresponding verb-noun action pair $a_k = (v_k, n_k)$ using a lightweight transformer-based classification head ${f}_{\text{AR}}$~\cite{zhao2024antgpt}, which is trained offline on the target dataset. The classification model minimizes the cross-entropy loss:
\begin{equation}
\mathcal{L}_{\text{AR}} = -\sum_{k=1}^{N} \left[ \log P(v_k |x_k) + \log P(n_k | {x}_k) \right],
\label{eq:ar_loss}
\end{equation}
where $v_k$ and $n_k$ are ground-truth labels. This action recognition process provides the history action sequence $\{a_1, a_2, \dots, a_N\}$, forming the foundational input for all subsequent reasoning stages.

\noindent \textbf{History Action Abstract Reasoning Module.}
This module reasons about the underlying task goal and environmental status by interpreting the historical actions. Given the history action sequence $\{a_1, a_2, \dots, a_k, \dots, a_N\}$ where each $a_k = (v_k, n_k)$, this module generates a History Action Abstract $\mathcal{I}_{\text{history}}$ in the form of natural language description that captures both the current object states and the inferred high-level human intention.
We formulate this as a conditional generation task: $\mathcal{I}_{\text{history}} = f_{\text{reason}}({a_1, \dots, a_N})$, where $f_{\text{reason}}$ is implemented using a large language model (LLM) with commonsense reasoning ability. Leveraging In-Context Learning (ICL), this module is provided with few-shot exemplars to bridge the semantic gap between low-level action sequences and high-level task goals. Specifically, these structured demonstrations guide the LLM to synthesize discrete actions into a coherent semantic narrative that describes the current environmental state and the most likely task goal. This process leverages the LLM's commonsense knowledge to infer latent intentions from observed interactions without requiring task-specific parameter updates.
As illustrated in Figure~\ref{fig:framework}, for a recognized sequence such as ${(\text{open}, \text{fridge}), (\text{take}, \text{cucumber}), (\text{take}, \text{onion})}$, the module generates $\mathcal{I}_{\text{history}}$: \textit{``With cucumbers and onions on hand, this person is about to make a salad."} This History Action Abstract $\mathcal{I}_{\text{history}}$ serves as essential semantic guidance for the subsequent Planner agent, ensuring that predicted future actions align with both the high-level human intention and the low-level environmental status.

\noindent\textbf{History Action Graph Construction Module.}
To maintain a structured memory of past interactions and object states, the Observer agent constructs and updates the History Action Graph $\mathcal{G} = (\mathcal{N}_{obs}, \mathcal{E})$ based on the recognized action sequence $\{a_1, a_2, \dots, a_N\}$. As illustrated in Figure~\ref{fig:framework}, this graph serves as a knowledge base that encodes both the temporal dependencies between actions and the evolving states of objects throughout the observed video.

\begin{itemize}
    \item \textbf{Nodes ($\mathcal{N}_{obs}$):} Each node $ n \in \mathcal{N}_{obs}$ corresponds to a unique object noun that has been interacted with in the observed video. Beyond static identification, each node maintains a dynamic attribute list $\text{Attr}(n)$ that records every action performed on the object along with precise start and end timestamps. Formally, for an object $n$, its attribute list contains tuples $(v, t_{\text{start}}, t_{\text{end}})$ for each action $a_k = (v, n)$ involving that object, enabling precise tracking of state changes over time.
    \item \textbf{Edges ($\mathcal{E}$):} Directed edges capture the temporal transition flow between consecutive actions. An edge $e_{ij} \in \mathcal{E}$ from action $a_i$ to action $a_j$ is established if $a_i$ and $a_j$ appear consecutively in the observed sequence, i.e., $j = i+1$. This structure preserves the chronological order of interactions.
    \item \textbf{Weights ($w_{ij}$):} Each edge $e_{ij}$ carries a weight $w_{ij}$ that quantifies the transition frequency between the two actions. This weight is incrementally updated each time the transition $a_i \rightarrow a_j$ is observed, effectively encoding a probabilistic prior over action transitions based on the observed history.
\end{itemize}

The History Action Graph $\mathcal{G}$ is constructed and updated as each new action $a_k$ is recognized by the Action Recognition module. Through this topological procedure, the Observer agent consolidates unstructured visual observations into a structured memory. This memory serves as a critical resource for the Verifier agent, providing explicit access to historical object states and action transition patterns that enable the verification of physical feasibility constraints as defined in Section~\ref{sec1:introduction}.

\subsection{Planner Agent}
\label{sec4-2:planner}
The Planner agent serves as the prediction unit. To addresses the second challenge outlined in Section~\ref{sec1:introduction}, the Planner aiming to generate draft of future actions $\hat{\mathcal{A}}_{fut} = \{\hat{a}_{N+1}, \dots, \hat{a}_{N+Z}\}$ that are consistent with the environmental context. 
Existing methods generate plans based predominantly on historical action narrations~\cite{kim2024palm} or inferred high-level goals~\cite{zhao2024antgpt}, thereby neglecting the current physical state of recent observations.

To mitigate this key limitation, the Planner synthesizes information from both the recognized historical action sequence $\{a_1, \dots, a_N\}$ and the History Action Abstract $\mathcal{I}_{\text{history}}$ with the current state description and the human intention. Furthermore, to handle the variable-length nature of real-world tasks, we design our training strategy to encompass varying observation ratios $\alpha \in \{25\%, 50\%, 75\%\}$ and correspondingly different future sequence lengths $Z$. This enables the Planner to adapt to diverse task durations rather than being constrained to predict a fixed number of actions. 
We then fine-tune a pre-trained Large Language Model (LLM) to serve as the action generation unit. To achieve parameter-efficient adaptation, we employ QLoRA, which introduces trainable low-rank adapters into the LLM's attention and feed-forward layers while keeping the base model weights frozen. 
This allows the model to learn task-specific temporal relationships between actions with efficient model training.
The model is trained to maximize the likelihood of the ground-truth future action sequence $\mathcal{A}_{fut}$ given the planner prompt $P_{\text{plan}}$:
\begin{equation}
\max_{\theta_{\text{QLoRA}}} \sum_{k=N+1}^{N+Z} \log P(a_k \mid P_{\text{plan}}, a_{N+1:k-1}; \theta_{\text{QLoRA}}),
\end{equation}
where $\theta_{\text{QLoRA}}$ represents the parameters of the low-rank adapters. 
This training objective enables the Planner to generate a draft action sequence $\hat{\mathcal{A}}_{fut}$ that is not only temporally consistent with the observed history and semantically aligned with the inferred human intention encoded in $\mathcal{I}_{\text{history}}$, but also adaptable to variable-length prediction scenarios. This draft is then passed to the Verifier for feasibility validation, completing the verification stage of our closed-loop framework.

\subsection{Verifier Agent}
\label{sec4-3:verifier}

To systematically assess feasibility, the Verifier ensures the three constraints defined in Section~\ref{sec3:problem_formulation}. 
Conditioned on the History Action Graph $\mathcal{G}$ from the Observer, the draft actions $\hat{a}_k$ generated by the Planner, the Verifier agent performs the feasibility check via a structured validation pipeline.
As illustrated in Figure~\ref{fig:framework}, the Verifier conducts a step-by-step feasibility check using Chain-of-Thought (CoT) reasoning through three phases: state reconstruction, constraint verification, and action refinement.

\noindent \textbf{State Reconstruction.} 
Before validating the draft action $\hat{a}_k$, the agent first parses the History Action Graph $\mathcal{G}$ to reconstruct the current environmental state $\text{Env}_N$ as defined in Eq.~(1). 
Leveraging few-shot exemplars, the model is guided to explicitly derive the current status of the objects from the history of actions.
Specifically, for each object node $n \in \mathcal{N}_{obs}$, the Verifier retrieves the attribute list $\text{Attr}(n)$ to identify the most recent interaction verb $v_{t^*(n)}$, where $t^*(n)$ represents the index of the last interaction. 
This process allows the agent to explicitly derive the object's physical status $\text{Post}(v_{t^*(n)})$, such as inferring the state ``opened'' from a past ``open'' action, thereby establishing the environment state $\text{Env}_N$ required for the subsequent consistency checks.

\noindent \textbf{Constraint Verification.}
As shown in Figure~\ref{fig:framework}, the Verifier conducts a step-by-step feasibility check to verify the constraints formulated in Section~\ref{sec3:problem_formulation}:
\begin{itemize}
    \item \textit{Object Availability ($\mathcal{C}_{obj}$):} 
    First, the Verifier implements the constraint in Eq.~(2) by confirming that the target object $\hat{n}_k$ exists within the graph's node set $\mathcal{N}_{obs}$, penalizing predictions that involves unobserved objects to reduce hallucinations.
    
    \item \textit{Affordance Validity ($\mathcal{C}_{aff}$):} 
    Second, the Verifier enforces Eq.~(3) by verifying whether the predicted verb $\hat{v}_k$ falls within the valid affordance set $\text{Aff}(\hat{n}_k)$, retrieving the interaction history recorded in $\text{Attr}(n)$.
    
    \item \textit{State Consistency ($\mathcal{C}_{sta}$):} 
    Finally, the Verifier evaluates the logical precondition constraint defined in Eq.~(4) by comparing the preconditions of the draft action, $\text{Pre}(\hat{v}_k)$, against the object's current state $\text{Post}(v_{t^*})$ derived in the State Reconstruction phase. This ensures $\text{Post}(v_{t^*}) \in \text{Pre}(\hat{v}_k)$.
    
\end{itemize}

To ensure robustness, few-shot exemplars guide the model to verify the above constraints against the draft actions. We provide specific violation examples of each constraint to enhance the model's discriminative ability in detecting infeasible actions.

\noindent\textbf{Action Refinement.}
Leveraging the results in Constraint Verification, the verifier employs In-Context Learning (ICL) to revise the infeasible action. The Verifier generates a revised action $\mathcal{A}^{*}_{fut}$ that prioritizes physical feasibility and is compliant with the constraints, effectively closing the ``Observe-Plan-Verify'' loop.

FactCheck addresses feasibility-aware long-term action anticipation with multi-agent collaboration via these three specialized agents. By closing the loop between action recognition, action prediction, and action verification, this collaborative framework ensures that the final action sequence $\mathcal{A}^{*}_{fut}$ is not only consistent with the human intention but also adheres to the feasibility constraints of the real-world environment.

\section{Experiments and Analysis}
\label{sec5}
In this section, we present a comprehensive empirical evaluation of our proposed FactCheck framework. 
First, we detail the experimental setup, including the mainstream LTA benchmarks, evaluation metrics, baseline methods, and implementation details. 
Subsequently, we report both the quantitative results and the qualitative results on two challenging ego-centric benchmarks: EPIC-Kitchens 55~\cite{damen2020epic} and EGTEA Gaze+~\cite{li2021eye}, comparing our approach with previous baseline methods. 
Finally, we conduct extensive ablation studies to validate the effectiveness of each modules and the collaborative mechanism.

\subsection{Experimental Setup}
\label{sec5-1:settings}

\subsubsection{Benchmarks and Metrics}

\noindent\textbf{EPIC-KITCHENS-55 (EK-55).}~\cite{damen2020epic}
EK-55 is a large-scale egocentric video dataset recording unscripted daily activities in native kitchen environments. It contains 55 hours of videos collected by 32 participants in 4 countries. The dataset is annotated with 39.6K action segments, covering 125 verbs and 352 nouns. We follow the standard train and test splits as EGO-TOPO~\cite{nagarajan2020ego} to evaluate action anticipation performance.

\noindent\textbf{EGTEA Gaze+(EGTEA).}~\cite{li2021eye}
EGTEA Gaze+ is another prominent egocentric video dataset with 86 unique videos of 32 subjects over 28 hours. The action annotations include 10325 instances of fine-grained actions covering 19 verbs and 53 nouns.  We follow the standard train and test splits as EGO-TOPO~\cite{nagarajan2020ego} to evaluate action anticipation performance.

\noindent\textbf{Evaluation Metrics.}
We follow the standard evaluation metrics for long-term action anticipation on both the EK-55 and the EGTEA datasets~\cite{nagarajan2020ego}. Given an observation ratio $\alpha$, the model observes the first $\alpha$ portion of a video and predicts the set of action classes that will occur in the remaining $(1-\alpha)$ portion. 
We evaluate across multiple anticipation horizons using $\alpha \in \{25\%, 50\%, 75\%\}$.
The primary evaluation metric is Mean Average Precision (mAP) across all classes.
To account for the long-tailed distribution of action classes in both datasets, we also report frequency-based metrics mAP-Freq, and mAP-Rare. We partition the class set $\mathcal{C}$ into frequent subsets $\mathcal{C}_{\text{freq}}$ and rare subsets $\mathcal{C}_{\text{rare}}$ as in EGO-TOPO~\cite{nagarajan2020ego}.
We report mAP-All, mAP-Freq, and mAP-Rare for each dataset to comprehensively evaluate the performance of FactCheck.

\subsubsection{Baselines}
To evaluate the effectiveness of FactCheck, we select the baseline methods that cover the major paradigms in the LTA field. We categorize the baseline methods into data-driven temporal modeling methods and LLM-driven agentic reasoning methods:

\noindent\textbf{(1) Data-driven temporal modeling methods.} This category focuses on learning the temporal patterns of the observed visual data to predict future actions through end-to-end training on large-scale datasets. Here are the baseline methods:
\begin{itemize}
    \item \textbf{I3D}~\cite{carreira2017quo}: A two-stream architecture that inflates pre-trained 2D image classification models into 3D ConvNets to learn seamless spatio-temporal feature extractors from video data.
    \item \textbf{ActionVLAD}~\cite{girdhar2017actionvlad}: A spatio-temporal aggregation architecture that pooling local convolutional features over space and time for end-to-end action classification.
    \item \textbf{Timeception}~\cite{hussein2019timeception}: A temporal convolutional layer-based method that models complex actions by using multi-scale temporal convolutions to capture long-range temporal dependencies and multi-scale temporal kernels to tolerate variations in temporal extents.
    \item 
    \textbf{VideoGraph}~\cite{hussein2019videograph}: A graph-based method that learns a soft undirected graph of latent concepts directly from video data to model the temporal structure inherent in minutes-long human activities.
    \item 
    \textbf{EGO-TOPO}~\cite{nagarajan2020ego}: A graph-based approach that constructs a topological map of the environment. It anticipates actions by reasoning about the functional affordances associated with the specific spatial zones visited by the agent.
    \item 
    \textbf{Anticipatr}~\cite{nawhal2022rethinking}: A two-stage learning approach employed to train a transformer-based model leverages both segment-level and video-level representations for long-term action anticipation.

\end{itemize}

\noindent \textbf{(2) LLM-driven agentic reasoning methods.} These methods leverage the semantic knowledge of pre-trained LLMs to infer future behaviors, typically utilizing a single-agent architecture. Here are the baseline methods:
\begin{itemize}
    \item 
    \textbf{AntGPT}~\cite{zhao2024antgpt}: An LLM-based action anticipation framework that infers the actor’s high-level goal from observations, and then generates future actions conditioned on the inferred goal, leveraging the LLM's language priors.
    \item 
    \textbf{PALM}~\cite{kim2024palm}: An LLM-based action anticipation framework that converts visual context into narrative captions via a VLM and leverages an LLM to generate future actions based on the narrative history.
\end{itemize}

\subsubsection{Implementation Details}
We detail the implementation settings for each module within the FactCheck framework. 

\textbf{Action Recognition.} 
Following AntGPT~\cite{zhao2024antgpt}, we first employ the CLIP-ViT-L/14-336px model to extract visual features without any model fine-tuning. For each video segment, we sample $T = 4$ frames at uniform temporal intervals. The lightweight transformer-based classification head follows the architecture of the action recognition module from AntGPT~\cite{zhao2024antgpt} for a fair comparison. The training objective is the cross-entropy loss over the verb and noun classes, as defined in Eq.~(\ref{eq:ar_loss}).

\textbf{History Action Abstract Reasoning.} This module leverages the commonsense reasoning capability of a large language model. We employ the Grok-4-1-fast-reasoning API~\cite{xai2025grok41fast} via in-context learning. The prompt template is constructed by concatenating the recognized history action sequence $\{a_1, \dots, a_N\}$ with a fixed instruction for the LLM to describe the current state of objects and the likely goal of the person in one sentence. The LLM response is then obtained as the History Action Abstract $\mathcal{I}_{\text{history}}$.

\textbf{Planner Agent Fine-tuning.} The Planner agent is implemented by fine-tuning pretrained Qwen2.5 models using the QLoRA technique. We train two model variants: a Qwen2.5-3B-Instruct model~\cite{qwen2.5} and a larger Qwen2.5-7B-Instruct model~\cite{qwen2.5}. For both variants, we construct a supervised fine-tuning dataset by formatting the input as a concatenation of the history actions and the History Action Abstract $\mathcal{I}_{\text{history}}$, with the target being the ground-truth future action sequence. The LoRA configuration uses a rank $r_{LoRA}=16$, alpha $\alpha_{LoRA}=32$, dropout rate $= 0.05$, and applies to all linear layers. 
For the Qwen2.5-3B-Instruct model, we fine-tune the model on a single NVIDIA RTX 4090 GPU, while the Qwen2.5-7B-Instruct model is fine-tuned across 8 NVIDIA RTX 3090 GPUs. We use bfloat16 mixed precision with 4-bit quantization for efficient model fine-tuning.

\textbf{Verifier Agent.} 
We implement the Verifier using the Grok-4-1-fast-reasoning API~\cite{xai2025grok41fast}. 
The History Action Graph $\mathcal{G}$ is linearized into a JSON Lines (JSONL) format.
We design a structured Chain-of-Thought (CoT) prompt to guide the agent through a three-step reasoning process:
(1) State Reconstruction, where the model interprets the linearized graph to reconstruct the current environmental state $\text{Env}_N$. 
Specifically, it parses the attribute list $\text{Attr}(u)$ of each object node to identify the most recent interaction $v_{t^*(n)}$ as defined in Eq.~1, thereby inferring the object's physical status;
(2) Constraint Verification, which executes a sequential checklist to validate the draft action against the graph data: 
The agent first checks \textit{Object Availability} by confirming that the target noun exists within the graph's node set $\mathcal{N}_{obs}$; 
Next, it assesses \textit{Affordance Validity} by verifying if the predicted verb is semantically compatible with the object's interaction history recorded in $\text{Attr}(u)$; 
Finally, it evaluates \textit{State Consistency} by comparing the logical preconditions of the draft action against the current object state $\text{Env}_N$ derived in step (1), such as rejecting an ``open'' action if the object is already in an ``opened'' state;
(3) Action Refinement.
Critically, the refinement stage is empowered by the In-Context Learning (ICL) mechanism. We embed $k=3$ manually crafted few-shot examples within the prompt, which demonstrate how to identify specific constraint violations and generate corresponding logic-based corrections. 
This enables the agent to adaptively revise infeasible predictions into verified actions that align with the history graph.

\subsection{Result Analysis}
\label{sec5-2:results}
We comprehensively evaluate the performance of FactCheck through both quantitative metrics in Section~\ref{sec5-2-1:quantitative} and qualitative case studies in Section~\ref{sec5-2-2:qualitative}. 

\subsubsection{Quantitative Analysis}
\label{sec5-2-1:quantitative}
To validate the effectiveness of our proposed framework, we evaluate our method on two mainstream LTA datasets: \textbf{EK-55} and \textbf{EGTEA Gaze+}.
Table~\ref{tab:ek55} presents the comparison on the \textbf{EK-55} validation set, where FactCheck consistently outperforms baselines in mAP scores across all metrics.
Table~\ref{tab:egtea} reports the results on the \textbf{EGTEA Gaze+} dataset. Our method achieves superior performance over existing methods, demonstrating the effectiveness of the FactCheck framework in diverse egocentric scenarios.

\begin{table}[h]
\centering
\caption{Comparison with existing methods on \textbf{EK-55} in mAP (\%). The first and second highest results are highlighted in \textbf{bold} and the best result in each column is marked with red background.}
\label{tab:ek55}
\resizebox{0.95\linewidth}{!}{
\begin{tabular}{l|ccc}
\toprule
Method & \textbf{ALL $\uparrow$} & \textbf{FREQ $\uparrow$} & \textbf{RARE $\uparrow$} \\ \midrule
I3D~\cite{carreira2017quo} & 32.7 & 53.3 & 23.0 \\
ActionVLAD~\cite{girdhar2017actionvlad} & 29.8 & 53.5 & 18.6 \\
Timeception~\cite{hussein2019timeception} & 35.6 & 55.9 & 26.1 \\
VideoGraph~\cite{hussein2019videograph} & 22.5 & 49.4 & 14.0 \\
EGO-TOPO~\cite{nagarajan2020ego} & 38.0 & 56.9 & 29.2 \\
Anticipatr~\cite{nawhal2022rethinking} & 39.1 & 58.1 & 29.1 \\ 
\midrule
AntGPT~\cite{zhao2024antgpt} & 40.1 & 58.8 & 31.9 \\
PALM~\cite{kim2024palm} & 40.4 & 59.3 & 30.3 \\
\midrule
\textbf{FactCheck-3B (Ours)} & \textbf{43.2} & \textbf{62.4} & \textbf{32.1} \\
\cellcolor{red!15}\textbf{FactCheck-7B (Ours)} & \cellcolor{red!15}\textbf{43.5} & \cellcolor{red!15}\textbf{63.3} & \cellcolor{red!15}\textbf{32.2} \\ 
\bottomrule
\end{tabular}
}
\end{table}

\begin{table}[h]
\centering
\caption{Comparison with existing methods on \textbf{EGTEA Gaze+} in mAP (\%). The first and second highest results are highlighted in \textbf{bold} and the best result in each column is marked with red background.}
\label{tab:egtea}
\resizebox{0.95\linewidth}{!}{
\begin{tabular}{l|ccc}
\toprule
Method & \textbf{ALL $\uparrow$} & \textbf{FREQ $\uparrow$} & \textbf{RARE $\uparrow$} \\ \midrule
I3D~\cite{carreira2017quo} & 72.1 & 79.3 & 53.3 \\
ActionVLAD~\cite{girdhar2017actionvlad} & 73.3 & 79.0 & 58.6 \\
Timeception~\cite{hussein2019timeception} & 74.1 & 79.7 & 59.7 \\
VideoGraph~\cite{hussein2019videograph} & 67.7 & 77.1 & 47.2 \\
EGO-TOPO~\cite{nagarajan2020ego} & 73.5 & 80.7 & 54.7 \\
Anticipatr~\cite{nawhal2022rethinking} & 76.8 & 83.3 & 55.1 \\ 
\midrule
AntGPT~\cite{zhao2024antgpt} & 80.2 & 84.8 & 72.9 \\
PALM~\cite{kim2024palm} & 80.7 & 85.0 & \textbf{73.5} \\
\midrule
\textbf{FactCheck-3B (Ours)} & \textbf{81.5} & \textbf{85.4} & 73.3 \\ 
\cellcolor{red!15}\textbf{FactCheck-7B (Ours)} & \cellcolor{red!15}\textbf{81.8} & \cellcolor{red!15}\textbf{86.9} & \cellcolor{red!15}\textbf{73.6} \\ \bottomrule
\end{tabular}
}
\end{table}

The quantitative results demonstrate the effectiveness of FactCheck across both datasets. 
As shown in Table~\ref{tab:ek55}, FactCheck-7B establishes a new SoTA with $43.5\%$ mAP-All, outperforming the previous best method PALM by a significant margin of \textbf{$3.1\%$} on EK-55. Notably, the improvements are consistent across both frequent ($+4.0\%$) and rare ($+1.9\%$) action classes.
FactCheck-3B also demonstrates competitive performance, achieving 43.2\% mAP-All and surpassing PALM by $2.8\%$, which highlights the efficiency of our architecture even with a smaller LLM backbone model.
As shown in Table~\ref{tab:egtea}, FactCheck-7B similarly achieve strong performance with $81.8\%$ mAP-All. FactCheck-7B surpasses PALM by $1.1\%$, with notable gains in the frequent category ($+1.9\%$).
Even the smaller FactCheck-3B outperforms the best baseline with $81.5\%$ mAP-All.
Overall, these experimental results indicate that our proposed Observe-Plan-Verify mechanism effectively ensures the physical feasibility of predictions, leading to robust improvements in both common and long-tail scenarios.

\begin{figure*}[htb]
    \centering
    \vspace{-8pt} 
    \includegraphics[width=0.8\linewidth]{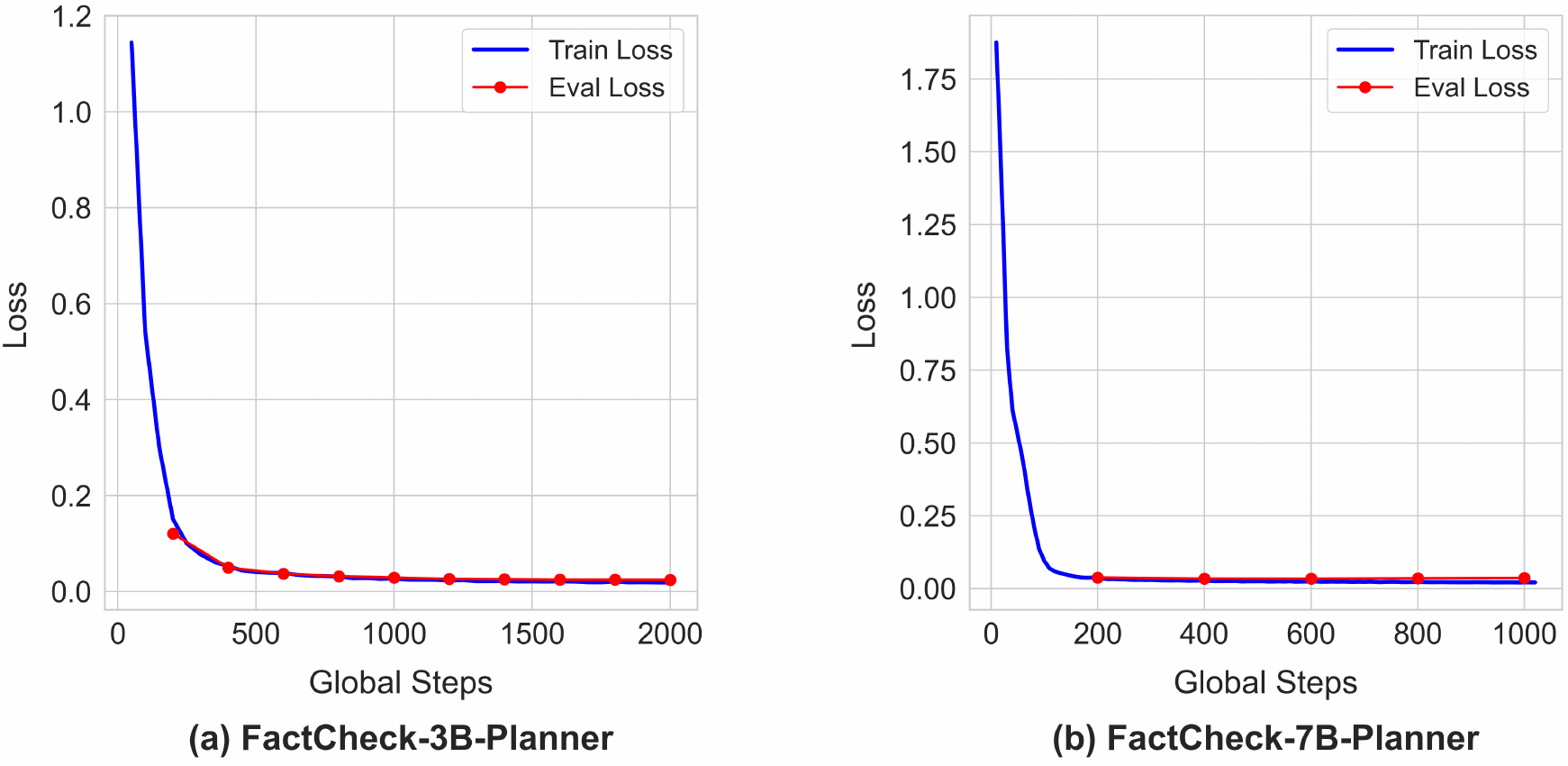}
    \vspace{-1mm}
    \caption{
    The training and validation loss of the Planner agent.
    }
    \label{fig:loss}
    \vspace{-5mm}
\end{figure*}

Traditional data-driven methods rely heavily on statistical patterns in training data and often fail to generalize to rare or unseen scenarios due to their limited capacity for semantic reasoning. While recent LLM-based single-agent approaches leverage commonsense knowledge for better intention inference, they operate in an open-loop manner without considering the current physical environment, leading to predictions that are semantically plausible but physically infeasible. In contrast, FactCheck's closed-loop "Observe-Plan-Verify" mechanism explicitly enforces physical feasibility constraints through structured environment modeling and iterative verification, resulting in more accurate and reliable long-term action anticipation. Both FactCheck-3B and FactCheck-7B achieve strong performance, demonstrating the effectiveness of our framework across different model scales. 
The slight but consistent advantage of FactCheck-7B can be attributed to its greater capacity for complex reasoning, which is further supported by our observation that the 7B model required significantly fewer training steps to achieve optimal performance. As illustrated in Figure~\ref{fig:loss}, both the 3B and 7B Planner models converge smoothly during training, and the stable validation loss confirms that neither model exhibits signs of overfitting.

\subsubsection{Qualitative Analysis}
\label{sec5-2-2:qualitative}
To further investigate the "Observe-Plan-Verify" mechanism of FactCheck, we present a qualitative case study on the "Continental Breakfast" task (video ID: P21-R04-ContinentalBreakfast) from the EGTEA Gaze+ dataset. This task contains a total of 79 actions. We set the observation ratio to 50\%, meaning the model observes the first 40 actions and predicts the subsequent sequence. 
In Table~\ref{tab:case_study}, we provide a case to demonstrate how the Verifier agent effectively corrects the Planner's infeasible draft actions.

As detailed in Table~\ref{tab:case_study}, the Observer agent first functions as the perception unit, utilizing the Action Recognition module to extract the history action sequence from the observed video frames. Based on these recognized actions, this agent generates a History Action Abstract via commonsense reasoning: \textit{``After putting on the pot, operating the stove, and inspecting the recipe, this person is going to make a continental breakfast.''} This abstract captures the high-level human intention and low-level environmental status. Furthermore, the Observer constructs a structured History Action Graph $\mathcal{G}$ that records the precise state of each object, for example, the \texttt{faucet} is [Off], and the \texttt{cabinet} is [Closed]. This graph serves as the structured memory for the subsequent verification process.

Conditioned on the History Action Abstract, the Planner agent generates a draft prediction sequence. While the high-level intention to continue preparing breakfast is correct, the Planner hallucinates two physically infeasible actions due to the lack of explicit physical verification in the open-loop generation process. Specifically, the Planner predicts \texttt{[Open cup]}, ignoring the object's operatable affordance, and \texttt{[Turn off faucet]}, disregarding the object's current state.

The Verifier agent then rigorously validates the draft against the History Action Graph $\mathcal{G}$. The Verifier first parses the graph to reconstruct the environment state. For the action \texttt{[Open cup]}, the Verifier queries the object attributes and detects a violation of Affordance Validity. It recognizes that a ``cup'' lacks a mechanically openable structure, making the action physically impossible. Leveraging In-Context Learning (ICL), the Verifier refines this action to \texttt{[Take cup]}. For the action \texttt{[Turn off faucet]}, the Verifier finds that the \texttt{faucet} is already in the [Off] state, as the last recorded action on the \texttt{faucet} is \texttt{Turn off}. Thus, the Verifier identifies a violation of state consistency and corrects the redundant action to \texttt{[Turn on faucet]}, which aligns with the task goal. This case validates that FactCheck successfully provides feasibility-aware long-term action anticipation.

\begin{table*}[t]
    \centering
    \caption{Qualitative example of the FactCheck correction process on the "Continental Breakfast" task. The red text highlights physically infeasible actions predicted by the Planner, while the green text shows the corrected actions after the Verifier's reasoning process.}
    \label{tab:case_study}
    \small 
    \begin{tabular}{p{2.5cm} p{0.8\textwidth}} 
        \toprule
        \textbf{Component} & \textbf{Action Sequence / Reasoning Process} \\
        \midrule
        \textbf{Ground Truth History Actions (50\%)} 
         & \texttt{... [Open cabinet] $\rightarrow$ [Close cabinet] $\rightarrow$ [Inspect/Read recipe] $\rightarrow$ [Turn on faucet] $\rightarrow$ [Turn off faucet] $\rightarrow$ [Put pot] $\rightarrow$ [Operate stove] $\rightarrow$ [Inspect/Read recipe]} \\
        \midrule
        \textbf{Observer} 
         & \textbf{History Action Recognition (50\%):} \texttt{... [Open cabinet] $\rightarrow$ [Close cabinet] $\rightarrow$ [Inspect/Read recipe] $\rightarrow$ [Turn on faucet] $\rightarrow$ [Turn off faucet] $\rightarrow$ [Put pot] $\rightarrow$ [Operate stove] $\rightarrow$ [Inspect/Read recipe]} \\
         & \textbf{History Action Abstract ($I_{history}$):} \textit{``After putting on the pot, operating the stove, and inspecting the recipe, this person is going to make a continental breakfast.''} \\
         & \textbf{History Action Graph ($\mathcal{G}$):} \\
         & \texttt{\small \{``nodes'': [\{``id'': ``cabinet'', ``state'': ``closed'', ``history'': [``Open'', ``Close'']\}, \{``id'': ``faucet'', ``state'': ``off'', ``history'': [``Turn on'', ``Turn off'']\}, \{``id'': ``pot'', ``state'': ``on\_stove'', ``history'': [``Put'']\}, \{``id'': ``stove'', ``state'': ``on'', ``history'': [``Operate'']\}, \{``id'': ``recipe'', ``state'': ``inspected'', ``history'': [``Inspect/Read'']\}], ``edges'': [\{``src'': ``cabinet'', ``tgt'': ``recipe'', ``w'': 1\}, ..., \{``src'': ``faucet'', ``tgt'': ``pot'', ``w'': 1\}, \{``src'': ``stove'', ``tgt'': ``recipe'', ``w'': 1\}]\}} \\
        \midrule
        \textbf{Planner}
                          & \textbf{Draft Prediction ($\hat{A}_{fut}$):} \texttt{[Open cabinet] $\rightarrow$ \textcolor{red}{[Open cup]} $\rightarrow$ [Close cabinet] $\rightarrow$ ... $\rightarrow$ [Take sponge] $\rightarrow$ [Put sponge] $\rightarrow$ \textcolor{red}{[Turn off faucet]} ...} \\
        \midrule
        \textbf{Verifier} & \textbf{CoT Reasoning :} \\
                          & \textit{1. State Reconstruction:} Parse graph $\mathcal{G}$. Current state of ``faucet'' is [Off] (last action: Turn off); ``cup'' is a container object. \\
                          & \textit{2. Constraint Verification:} \\
                          & \quad - Action [Open cabinet]: Valid. \\
                          & \quad - Action [Open cup]: \textbf{Violation: Affordance Validity Error}. ``Cup'' lacks a mechanically openable structure. \\
                          & \quad - ... \\
                          & \quad - Action [Turn off faucet]: \textbf{Violation: State Consistency Error}. Current state is already [Off]; turning it off again is redundant and physically infeasible. \\
                          & \textit{3. Action Refinement:} \\
                          & \quad - Replace ``Open cup'' with ``Take cup'' (correct interaction). \\
                          & \quad - Replace ``Turn off faucet'' with ``Turn on faucet'' (logical next step for washing). \\
        \midrule
        \textbf{Final Output} & \texttt{[Open cabinet] $\rightarrow$ \textcolor{green}{[Take cup]} $\rightarrow$ [Close cabinet] $\rightarrow$ ... $\rightarrow$ [Take sponge] $\rightarrow$ [Put sponge] $\rightarrow$ \textcolor{green}{[Turn on faucet]} $\rightarrow$ ... } \\
        \midrule
        \textbf{Ground Truth Future Actions} & \texttt{[Open cabinet] $\rightarrow$ [Take cup] $\rightarrow$ [Close cabinet] $\rightarrow$ ... $\rightarrow$ [Take sponge] $\rightarrow$ [Put sponge] $\rightarrow$ [Turn on faucet] $\rightarrow$ ... } \\
        \bottomrule
    \end{tabular}
\end{table*}


\subsection{Ablation Study}
\label{sec5-3:ablation}
To comprehensively evaluate the contribution of our FactCheck framework, we conduct ablation studies from three perspectives: (1) the effectiveness of each individual module within the FactCheck framework, (2) the generalizability across different large language model backbones, and (3) the impact of the multi-agent architecture compared to single-agent baselines. These experiments are performed on the EGTEA Gaze+ dataset.

\subsubsection{Effectiveness of each component}

To validate the effectiveness of our proposed FactCheck, we conduct an ablation study on the EGTEA Gaze+ dataset, as summarized in Table~\ref{tab:ablation_collab}. 
We compare the full FactCheck-7B model against three variants, each removing a key module: the History Action Abstract, the History Action Graph, and the Verifier.

\begin{table}[h]
\centering
\caption{Ablation study on different components of FactCheck on the \textbf{EGTEA Gaze+} dataset in mAP (\%).}
\label{tab:ablation_collab}
\begin{tabular}{>{\raggedright\arraybackslash}p{3.5cm}|ccc}
\toprule
\textbf{Variant} & \textbf{ALL $\uparrow$} & \textbf{FREQ $\uparrow$} & \textbf{RARE $\uparrow$} \\
\midrule
\textbf{FactCheck-7B} & \cellcolor{red!15}\textbf{81.8} & \cellcolor{red!15}\textbf{86.9} & 73.6  \\
w/o History Action Abstract& 81.5 & 86.4 & 73.2\\
w/o History Action Graph & 81.1 & 85.6 & 74.1 \\
w/o Verifier & 81.0 & 84.9 & \cellcolor{red!15}\textbf{74.4}  \\
\bottomrule
\end{tabular}
\end{table}

\textbf{Impact of the History Action Abstract.} 
Removing the History Action Abstract leads to a consistent performance drop across all metrics. 
This indicates that the summarization of the current environmental state and the high-level human intention serve as a crucial global context, helping the model maintain a memory of the historical actions. The History Action Abstract benefits both frequent and rare actions, suggesting that this module helps enhance the overall task performance.

\textbf{Impact of the History Action Graph.} 
The removal of the structured graph results in a decline in mAP-All and mAP-Freq. 
This confirms that explicitly modeling noun-verb relationships via a graph is essential for capturing the historical actions. 
However, we observe an increase in mAP-Rare. This suggests that while the graph effectively enhances the task performance via action verification, it may introduce a bias towards common transitions, potentially suppressing some less common actions in the long tail distribution.

\textbf{Impact of the Verifier.} 
Removing the Verifier also leads to a drop in mAP-All and mAP-Freq performance. 
This result underscores the critical role of the "Observe-Plan-Verify" mechanism. 
Without the feasibility constraints imposed by the Verifier, the model is prone to hallucinating physically infeasible actions, resulting in performance degradation.
We observe that the FactCheck w/o Verifier achieves a higher score on Rare action classes. 
This suggests a trade-off where the Verifier acts as a strong critic that prioritizes physical feasibility over risky long-tail predictions.

In conclusion, the full model achieves the best balance, yielding the highest mAP-All and mAP-Freq scores, validating that each module is indispensable for the LTA task.


\subsubsection{Generalizability across large language model backbones}

To evaluate the generalizability of the FactCheck framework, we investigate the performance across different large language model backbones. To ensure a fair comparison, we replace the default grok-4-1-fast-reasoning backbone used in the Observer and Verifier agents with gemini-3-flash-preview, while keeping all other components strictly identical. 
As reported in Table \ref{tab:generalizability}, FactCheck maintains robust performance across different backbones. The variant employing gemini-3-flash-preview achieves an mAP-All of 81.3\%, which is highly competitive with the grok-4-1-fast-reasoning and still significantly outperforms existing state-of-the-art LTA methods. This ablation study demonstrates that our "Observe-Plan-Verify" mechanism consistently enhances action feasibility by leveraging the inherent reasoning capabilities of various LLMs. This suggests that the effectiveness of FactCheck stems from its collaborative architecture rather than a specific choice of the LLM.

\begin{table}[h]
\centering
\caption{Generalizability of FactCheck across different LLM backbones (Observer \& Verifier) on \textbf{EGTEA Gaze+}.}
\label{tab:generalizability}
\begin{tabular}{p{4cm}|ccc}
\toprule
\textbf{Backbone (Observer \& Verifier)} & \textbf{ALL $\uparrow$} & \textbf{FREQ $\uparrow$} & \textbf{RARE $\uparrow$} \\
\midrule
gemini-3-flash-preview & 81.3 & \cellcolor{red!15}\textbf{85.5} & 73.0 \\
\textbf{grok-4-1-fast-reasoning (Default)} & \cellcolor{red!15}\textbf{81.5} & 85.4 & \cellcolor{red!15}\textbf{73.3} \\
\bottomrule
\end{tabular}
\end{table}

\subsubsection{Impact of multi-agent architecture}

To validate the necessity of our multi-agent collaboration framework, we compare FactCheck against single-agent baselines. All single-agent baselines are provided with the same input information: the sequence of history actions and the History Action Abstract. 
We design three specific baselines to evaluate the backbone models without the multi-agent architecture: (1) Qwen-2.5-3B-Instruct (LoRA): We fine-tune a single Qwen-2.5-3B-Instruct model using LoRA. This investigates whether the performance gains can be achieved simply by fine-tuning a single LLM agent rather than through multi-agent collaboration. (2) Grok-4-1-fast-reasoning (Zero-shot): A single-agent baseline using the Grok-4-1-fast-reasoning model without task-specific demonstrations. This evaluates the backbone model's inherent reasoning capabilities on the LTA task. (3) \textbf{Grok-4-1-fast-reasoning (Few-shot)}: A single-agent baseline using the Grok-4-1-fast-reasoning model with few-shot guidance. This setup tests whether a state-of-the-art LLM agent matches the FactCheck framework's performance without its collaborative architecture.
As shown in Table~\ref{tab:single_vs_multi_agent}, FactCheck-3B consistently outperforms all single-agent baselines across all metrics. Specifically, compared to the Qwen-2.5-3B-Instruct (LoRA) baseline, our framework achieves a substantial improvement of 6.2\% in mAP-All. 
Furthermore, FactCheck-3B surpasses the Grok-4-1-fast-reasoning (Zero-shot) baseline by 3.3\% and the Grok-4-1-fast-reasoning (Few-shot) baseline by 1.9\% in mAP-All. We observe that single agents often hallucinate non-existent objects, violate object affordances, or disregard object states, as they operate in an open-loop manner without a mechanism to explicitly verify the physical feasibility of their predictions against the environment. This ablation study shows our multi-agent collaboration mechanism effectively unlocks the potential of LLMs for feasibility-aware LTA.

\begin{table}[h]
\centering
\caption{Performance comparison between single-agent baselines and the multi-agent framework FactCheck on the \textbf{EGTEA Gaze+} dataset in mAP (\%).}
\label{tab:single_vs_multi_agent}
\begin{tabular}{>{\raggedright\arraybackslash}p{4cm}|ccc}
\toprule
\textbf{Method} & \textbf{ALL $\uparrow$} & \textbf{FREQ $\uparrow$} & \textbf{RARE $\uparrow$} \\
\midrule
Qwen-2.5-3B-Instruct (LoRA) & 75.3 & 82.1 & 56.8 \\
Grok-4-1-fast-reasoning (Zero-shot) & 78.2 & 80.6 & 71.4 \\
Grok-4-1-fast-reasoning (Few-shot) & 79.6 & 82.9 & 73.1 \\
\midrule
\textbf{FactCheck-3B (Ours)} & \cellcolor{red!15}\textbf{81.5} & \cellcolor{red!15}\textbf{85.4} & \cellcolor{red!15}\textbf{73.3} \\
\bottomrule
\end{tabular}
\end{table} 
\vspace{-0.5cm}

\subsection{Future Works}

In this paper, we propose FactCheck, a multi-agent framework that enables feasibility-aware Long-term Action Anticipation (LTA). Feasibility-aware LTA serves as a critical capability for the advancement of embodied intelligence. We discuss future research directions from the perspectives of long-horizon robot manipulation, proactive human-robot collaboration (HRC) and handling unseen objects in an open-world environment below:

\textbf{Long-horizon Robot Manipulation.} Long-horizon robot manipulation is a cornerstone of autonomous systems, enabling agents to perform complex, multi-step tasks. However, it remains a significant challenge for LLM-based robot manipulation methods due to the compounding uncertainty of long-term future sequences and the strict physical constraints of real-world environments, where a single error may disrupt an entire workflow. FactCheck addresses this by generating action sequences that are verified for physical feasibility rather than just semantic coherence. These validated sequences can be executed by a robot using a library of low-level motion primitives, where each step in the plan is directly mapped to a specific control policy. By bridging the gap between high-level planning and physical execution, our framework enables robots to reliably accomplish complex real-world tasks, including household activities such as cooking and cleaning.

\textbf{Proactive Human-Robot Collaboration (HRC).} FactCheck empowers Proactive HRC by enabling service robots to accurately anticipate human intentions and provide timely assistance without the need for human explicit instructions. In collaborative environments such as a kitchen, if our FactCheck framework predicts a user intends to "cut a cucumber," the robot can proactively retrieve and hand over a knife. Crucially, the "feasibility-aware" nature of our model ensures the safety and reliability of these interactions. Unlike open-loop approaches that may hallucinate unnecessary or impossible steps, FactCheck explicitly verifies that the required tool is available and the target object is in the appropriate state. This validation prevents the robot from performing confusing or unsafe actions based on false predictions, thereby fostering reliability in human-robot collaboration.

\textbf{Handling Unseen Objects in Open-World Environments.} While our current framework applies Object Availability constraints based on observed history to effectively mitigate object hallucination, this strict constraint may unintentionally filter out correct predictions involving objects that have not yet appeared in the video, such as when entering a new room or opening an unseen cabinet. In future work, we plan to incorporate open-vocabulary object detection to estimate the likelihood of unobserved objects entering the scene. 
This would enable a better balance between feasibility verification and generalization to open-world scenarios.

\section{Conclusions}
\label{sec6:conclusion}

In this paper, we present FactCheck, a multi-agent collaboration framework for long-term action anticipation that enhances the feasibility of predicted actions in a closed-loop “Observe-Plan-Verify” manner. The proposed Observer agent encodes the video observation into a dual-form memory of historical actions, comprising a concise textual description and a structured graph. The Planner agent anticipates future actions consistent with the textual action memory. The Verifier agent explicitly verifies future actions with the graph-based memory, enabling physically executable actions. Factcheck achieves effective performance on the EK-55 and EGTEA Gaze+ benchmark and outperforms other SoTA baselines. Ablation studies demonstrate the necessity of each component in our multi-agent framework. Our work establishes a new paradigm for feasibility-aware long-term action anticipation, effectively closing the loop of action recognition, action prediction, and action verification.

\bmsection*{Acknowledgments}
This work was supported by HK RGC General Research Fund (No. PolyU-15235424), “Research on Key Technologies for Systematic Artificial Intelligence Agents” project under China Mobile Innovation and Research Institute (No. R24114H7), PolyU Internal Research Fund (No. BDZ3), PolyU LTC Project (No. TDLEG25-28/IICA/P/05) and Research Institute for Artificial Intelligence of Things, The Hong Kong Polytechnic University.

\bibliography{0-reference}

\end{document}